\documentclass[10pt,twocolumn,twoside]{IEEEtran}
\usepackage[colorlinks=true, urlcolor=blue, linkcolor=red]{hyperref}
\usepackage{booktabs} 
\usepackage{amsmath}
\usepackage{algorithm}
\usepackage{graphicx}
\usepackage{float}
\usepackage{makecell}
\usepackage{bbding}
\usepackage{calc,amsfonts,amssymb,amsmath,bm,url,color,theorem,graphicx,cite}
\usepackage{psfrag,subfigure,float}
\usepackage{multirow,subfigure,amsmath,epsfig,amsfonts,amssymb,psfig,graphics,psfrag,theorem,calc,subfigure,url,bm,cite}
\usepackage{stfloats,hyperref}  
\usepackage{algorithm,algorithmic}
\usepackage{diagbox} 
\usepackage{hhline}
\usepackage{xcolor}
\usepackage{booktabs}
\usepackage{makecell}
\usepackage{colortbl}
\usepackage{xcolor} 

\definecolor{bestColor}{RGB}{255, 0, 0}    
\definecolor{secondBestColor}{RGB}{0, 0, 255} 
\definecolor{thirdBestColor}{RGB}{0, 100, 0}  

\newcommand{\best}[1]{\textcolor{bestColor}{\textbf{#1}}}      
\newcommand{\secondBest}[1]{\textcolor{secondBestColor}{\textbf{#1}}} 

\definecolor{best}{rgb}{1,0,0}      
\definecolor{secondbest}{rgb}{0,0,1}  

\usepackage{mathtools}

\makeatletter
\def\multilimits@{\bgroup
	\Let@
	\restore@math@cr
	\default@tag
	\baselineskip\fontdimen10 \scriptfont\tw@
	\advance\baselineskip\fontdimen12 \scriptfont\tw@
	\lineskip\thr@@\fontdimen8 \scriptfont\thr@@
	\lineskiplimit\lineskip
	\vbox\bgroup\ialign\bgroup\hfil$\m@th\scriptstyle{##}$\hfil\crcr}
\def\Sb{_\multilimits@}
\def\endSb{\crcr\egroup\egroup\egroup}
\makeatother

{\begin{list}{}%
		{\setlength{\rightmargin}{0pc}%
			\setlength{\leftmargin}{1pc} }} %
	{\end{list}}

\newlength{\twidth}
\definecolor{orange}{RGB}{255,107,0}

\theorembodyfont{\rmfamily}

{\begin{list}{}{
			\settowidth{\labelwidth}{\mbox{\textnormal{#1}}}%
			\setlength{\leftmargin}{\labelwidth+\labelsep}}}%
	{\end{list}}

\newcommand\bA{\ensuremath{{\bm A}}}
\newcommand\bB{\ensuremath{{\bm B}}}
\newcommand\bC{\ensuremath{{\bm C}}}
\newcommand\bD{\ensuremath{{\bm D}}}

\newcommand\bK{\ensuremath{{\bm K}}}

\newcommand\bN{\ensuremath{{\bm N}}}
\newcommand\bO{\ensuremath{{\bm O}}}

\newcommand\bQ{\ensuremath{{\bm Q}}}

\newcommand\bV{\ensuremath{{\bm V}}}
\newcommand\bW{\ensuremath{{\bm W}}}
\newcommand\bX{\ensuremath{{\bm X}}}
\newcommand\bY{\ensuremath{{\bm Y}}}
\newcommand\bZ{\ensuremath{{\bm Z}}}

\definecolor{orange}{RGB}{255,107,0}

\author{Chih-Chung Hsu,~\IEEEmembership{Senior Member,~IEEE}, \\
        Chih-Chien~\!Ni, Chia-Ming~\!Lee, and Li-Wei Kang,~\IEEEmembership{Member,~IEEE}}
\title{CSAKD: Knowledge Distillation with Cross Self-Attention for Hyperspectral and Multispectral Image Fusion 
\thanks{This study was supported partly by National Science and Technology Council (NSTC), Taiwan, under Grants NSTC 110-2636-E-006-026, 110-2222-E-006-012, 111-2634-F-007-002, 110-2218-E-006-026, and 111-2221-E-003-019-MY3.}
\thanks{\textit{(Corresponding author: Li-Wei Kang.)}}
\thanks{C.-C. Hsu and C.-M. Lee are with the Institute of Data Science, and with Miin Wu School of Computing, National Cheng Kung University, Tainan, Taiwan (R.O.C.)  
(e-mail: cchsu@gs.ncku.edu.tw, zuw408421476@gmail.com).}
\thanks{C.-C. Ni and L.-W Kang are with the Department of Electrical Engineering, National Taiwan Normal University, Taipei, Taiwan (R.O.C.) 
(e-mail: ken0934032237@gmail.com, lwkang@ntnu.edu.tw).}
}

\begin{document}

\maketitle

\begin{abstract}
Hyperspectral imaging, capturing detailed spectral information for each pixel, is pivotal in diverse scientific and industrial applications. Yet, the acquisition of high-resolution (HR) hyperspectral images (HSIs) often needs to be addressed due to the hardware limitations of existing imaging systems. A prevalent workaround involves capturing both a high-resolution multispectral image (HR-MSI) and a low-resolution (LR) HSI, subsequently fusing them to yield the desired HR-HSI. Although deep learning-based methods have shown promising in HR-MSI/LR-HSI fusion and LR-HSI super-resolution (SR), their substantial model complexities hinder deployment on resource-constrained imaging devices.
This paper introduces a novel knowledge distillation (KD) framework for HR-MSI/LR-HSI fusion to achieve SR of LR-HSI. Our KD framework integrates the proposed Cross-Layer Residual Aggregation (CLRA) block to enhance efficiency for constructing Dual Two-Streamed (DTS) network structure, designed to extract joint and distinct features from LR-HSI and HR-MSI simultaneously. To fully exploit the spatial and spectral feature representations of LR-HSI and HR-MSI, we propose a novel Cross Self-Attention (CSA) fusion module to adaptively fuse those features to improve the spatial and spectral quality of the reconstructed HR-HSI. Finally, the proposed KD-based joint loss function is employed to co-train the teacher and student networks. 
Our experimental results demonstrate that the student model not only achieves comparable or superior LR-HSI SR performance but also significantly reduces the model-size and computational requirements. This marks a substantial advancement over existing state-of-the-art methods. The source code is available at \hyperlink{https://github.com/ming053l/CSAKD}{https://github.com/ming053l/CSAKD}.
\end{abstract}


\section{Introduction}

\begin{figure}
    \begin{center}
    \includegraphics[width=0.5\textwidth] {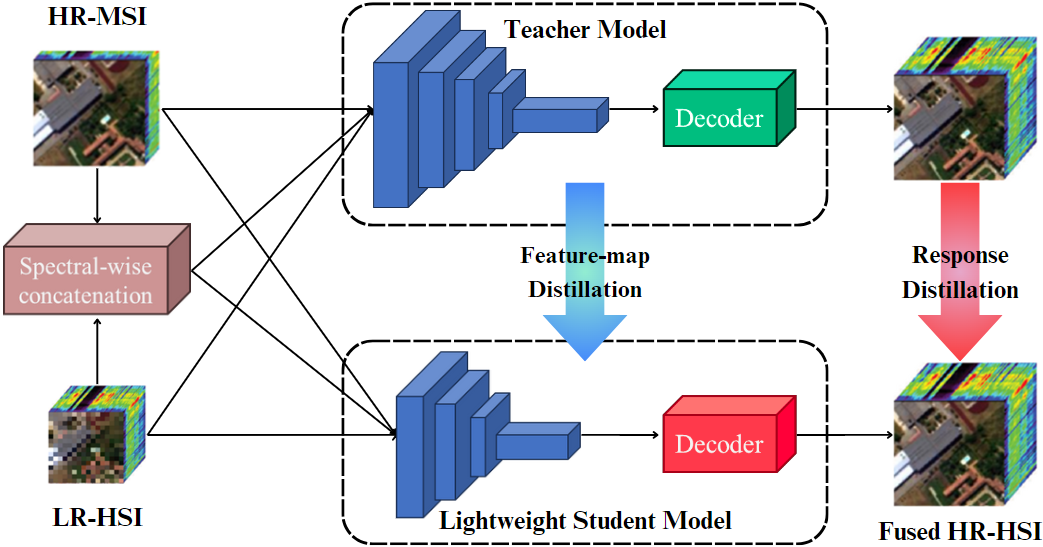}
    \end{center}
    \caption{{The brief illustration of proposed CSAKD framework by adaptively fusing the features of the LR-HSI and HR-MSI.}}
    \label{fig:res_header}
\end{figure}

Hyperspectral imaging aims to capture information based on dense spectral sensing at each image pixel of a scene. Compared with conventional imaging modalities, hyperspectral images (HSIs) include a wider spectral range, with the number of channels ranging from ten to hundreds. HSIs have been shown to enable a wide range of applications in the fields of industry, science, military, agriculture, and medicine \cite{1}. However, the extreme limitation of hardware of hyperspectral image sensing systems in the miniaturized satellite often restrict that the spectral or spatial resolution could not be large enough. In practice, the general solution is to capture the image of high spatial resolution together with limited spectral bands. That is, existing sensing systems usually capture the high-resolution (HR) multispectral images (MSIs), i.e., HR-MSIs, and the low-resolution (LR) HSIs, i.e., LR-HSIs. To further enhance the spatial resolution of LR-HSI, super-resolution (SR) of LR-HSI achieved by fusing HR-MSI and LR-HSI to obtain the corresponding HR-HSI has been a promising way \cite{2,3,4} in recent research direction.

Several traditional fusion methods have been presented with the development of LR-HSI and HR-MSI fusion techniques (e.g., \cite{5,6}). For example, sparse representation-based \cite{7}, and low-rank-based \cite{8} matrix decomposition-guided fusion frameworks were proposed to achieve feasible performance for the SR of LR-HSI. 
Benefiting the advantages of recent deep learning (DL) related techniques, such as image restoration \cite{9,10}, image classification \cite{11,12}, and object detection \cite{13,14}, DL-based HR-MSI and LR-HSI fusion methods have been proposed recently for obtaining the better spectral and spatial quality of the reconstructed HR-HSI \cite{20,21,22,23,24,25,26,27,28,29}. However, the current state-of-the-art DL-based HR-MSI/LR-HSI fusion methods may still suffer from higher model complexity or insufficient image detail reconstruction due to the lack of fully exploiting the spectral and spatial feature representation from both HR-MSI and LR-HSI.

To design a lightweight deep HR-MSI/LR-HSI fusion model and produce sufficiently good HR-HSI of the input LR-HSI, in this paper, we propose a knowledge distillation (KD)-based LR-HSI and HR-MSI fusion method to meet the massive requirements of the real-time applications, as the power supply in the world is becoming emerged. In the proposed framework, we first train a sophisticated teacher network with excellent HR-MSI/LR-HSI fusion performance. Then, we distill the knowledge from the teacher network into a lightweight student network to achieve high-quality outcomes in both the spectral and spatial domains.
To effectively guide student network learning, the KD-loss is adopted to ensure higher similarity between the feature maps, respectively, generated from the teacher and the student networks based on the response-based KD approach \cite{15}, thereby improving the performance of the student network. Since the oversimplified student network could be harmful to the quality of the reconstructed HR-HSI, a good and simple network architecture is essential. Moreover, it is well-known that the feature representation of HR-MSI and LR-HSI could be significantly different from each other, implying that directly fusing them without dynamically determining the corresponding weights could result in restricted performance. Therefore, to fully exploit the spatial and spectral features of LR-HSI and HR-MSI without increasing the parameters of the teacher/student networks, we propose a novel Dual Two-Streamed (DTS) network based on our Cross-Layer Residual Aggregation (CLRA) block with the Cross-Self-Attention (CSA) fusion module for judiciously extracting the needed spatial and spectral information for obtaining the better quality of the reconstructed HR-HSI, as illustrated in Fig. \ref{fig:res_header}. In this way, the proposed DTS Network not only achieves state-of-the-art performance but also reduces the computational and space complexities simultaneously. The major novelties and contributions of this paper are three-fold:

\begin{itemize}
    \item{to the best of our knowledge, we are among the first to propose a response-based KD framework to learn a lightweight HR-MSI and LR-HSI fusion model;}
    \item{the proposed DTS Network effectively incorporates the spatial and spectral features from LR-HSI/HR-MSI dynamically by using our CSA fusion module; and}
    \item{the proposed method has been shown to outperform several state-of-the-art LR-HSI/HR-MSI fusion models in terms of different metrics.}
    
\end{itemize}
	
The rest of this paper is organized as follows. In Sec. II, we briefly introduce the related works, including traditional frameworks for SR of LR-HSI, the DL-based frameworks for SR of LR-HSI, and related KD techniques. In Sec. III, we present the proposed DTS Network with KD framework for learning a lightweight deep HR-MSI/LR-HSI fusion network. In Sec. IV, experimental results, and ablation studies are demonstrated. Finally, Sec. V concludes this paper.

\section{Related Works}
This section provides an overview of the methodologies developed for enhancing the spatial and spectral resolution of hyperspectral images. The evolution of these methodologies spans from traditional techniques, leveraging sparse representations and low-rank matrix factorizations, to contemporary DL-based approaches that exploit the representational power of convolutional neural networks (CNNs) for superior fusion outcomes. Additionally, we discuss the emergent strategy of the KD aimed at refining model efficiency and facilitating deployment on resource-constrained devices.

\subsection{Optimization-based Approach}
In \cite{16}, a pioneer MSI Pan-sharpening framework was presented, where the goal is to fuse LR-MSI and HR-panchromatic image (with single band and high spatial resolution) of the same scene to generate an image with high spectral and spatial resolutions. Moreover, based on the sparse or low-rank image prior knowledge of HSIs, several sparse representation-based or low-rank-based image fusion frameworks were presented for the SR of HSI or HR-MSI/LR-HSI fusion \cite{7,8,17,18,19}. For example, in \cite{7}, an SR method for LR-HSI was proposed, where the prediction of the HR-HSI is formulated as a joint derivation task of the HSI dictionary and the sparse codes relying on the spatial-spectral sparsity of HSIs. In addition, a group spectral embedding-based HR-MSI/LR-HSI fusion method was presented in \cite{8}, where the manifold structures of spectral bands and the low-rank structure of HR-HSIs were explored. A spatial and spectral fusion model was also proposed in \cite{17} by using sparse matrix factorization to fuse remote sensing images of HR with low spectral resolution (similar to HR-MSI) and LR with high spectral resolution (similar to LR-HSI). An image fusion framework relying on spectral unmixing and sparse coding was similarly proposed in \cite{18} to fuse HR-MSI and LR-HSI. Furthermore, a coupled sparse tensor factorization framework was presented in \cite{19} for fusing HR-MSI and LR-HSI, where estimating the dictionaries and core tensor was formulated as a coupled tensor factorization problem. Since these traditional methods rely on some image priors, such as sparse or low-rank, some real-world scenarios not fitting these assumptions may introduce some performance degradation. While the optimization-based approach usually requires high-precision computation, and hard to deploy those algorithms into moderate AI-chip since it is hard to parallelize (e.g., eigendecomposition is often used in optimization-based methods), reducing the computational complexity with promising performance is highly desired.

\subsection{Deep Learning-based Approach}
DL-based strategy has shown promise in HR-MSI/LR-HSI fusion tasks. 
With the development of DL technology, such as the powerful representation learning ability of CNNs, several SR frameworks for LR-HSI or the fusion of HR-MSI and LR-HSI have been recently proposed. A 3-D CNN was used in \cite{20} to fuse multispectral and hyperspectral images to generate an HR-HSI, where the dimensionality of the HSI was reduced prior to the fusion process to significantly reduce the computational complexity. A blind HR-MSI/LR-HSI fusion problem was formulated and solved based on DL in \cite{21}, where the estimation of the observation model and fusion process are optimized iteratively and alternatively during the SR reconstruction. In addition, an HSI reconstruction algorithm with a data-driven prior relying on an optimization-inspired DL was presented in \cite{22}, where the prior was learned based on both the local coherence and dynamic characteristics of HSIs. Moreover, an end-to-end DL network was proposed in \cite{23} to jointly learn multi-scale spatial-spectral features for HR-MSI and LR-HSI fusion (denoted by MSSJFL). In addition, a lightweight deep model-based progressive zero-centric residual network (denoted by PZRes-Net) was presented in \cite{24} for SR of HSI, where the spectral-spatial separable convolution operations with dense connections were used to efficiently learn the residual image. In \cite{25}, a dual-UNet-based architecture with a multi-stage details injection strategy was presented for fusing HR-MSI and LR-HSI, where a multi-scale spatial-spectral attention module was utilized. Furthermore, a deep hyperspectral image fusion network (denoted by DHIF-Net) was proposed in \cite{26}, where an end-to-end optimization strategy of iterative spatial-spectral regularization was implemented. On the other hand, an unregistered and unsupervised mutual Dirichlet-Net was presented in \cite{27} for SR of HSI. An Interpretable deep neural network designed for HR-MSI/LR-HSI fusion was proposed in \cite{28}. An interpretable deep model named by spatial–spectral dual-optimization model-driven deep network was also presented in \cite{29} for HR-MSI/LR-HSI fusion. 

However, considering that the lightweight models designed manually could be tedious and, thus, hard to guarantee their performance, we propose an effective network architecture (i.e., DTS Network) to ensure a promised performance and followed by applying the KD-based approach to reduce the computational and spatial complexity without significant performance degradation. 

\subsection{Knowledge Distillation}
Directly deploying a sophisticated network into low-power devices is infeasible due to its extreme limitation of memory and computational resources. KD manner offers a solution by training efficient "student" models guided by complex "teacher" networks, aiming for the student to match or exceed the teacher's performance. This process involves strategic knowledge transfer, which can be categorized into response-based \cite{30}, feature-based \cite{31}, and relation-based \cite{32} KD schemes. 

Response-based KD focuses on emulating the teacher model's final output, enabling the student model to learn directly from these predictions, as seen in \cite{30}. Feature-based KD expands on this by using outputs from both the final and intermediate layers of the teacher model, enriching the student's learning with deeper insights, exemplified by \cite{31}. Relation-based KD, on the other hand, transfers inter-layer relationships to provide a nuanced understanding of model behaviors, as detailed in \cite{32}.

The application of KD in HSI processing tasks, including segmentation and pan-sharpening \cite{33, 34}, showcases its potential for enhancing HSI and MSI fusion with lower computational and space complexity. Directing adopts response-based KD, which could be enough to distill the knowledge in the teacher network to that in the student one. In this paper, we would like to emphasize that the efficient and effective network architecture for teacher and student models is essential, while KD is a way to further reduce the complexity by knowledge transfer. Therefore, we do not focus on the selection of the KD framework in this study. 

\section{Proposed Dual Two-Streamed Network via Cross-Self-Attention Fusion}\label{sec:theory}
\begin{figure*}[htbp]
\centering
\includegraphics[width=0.95\linewidth]{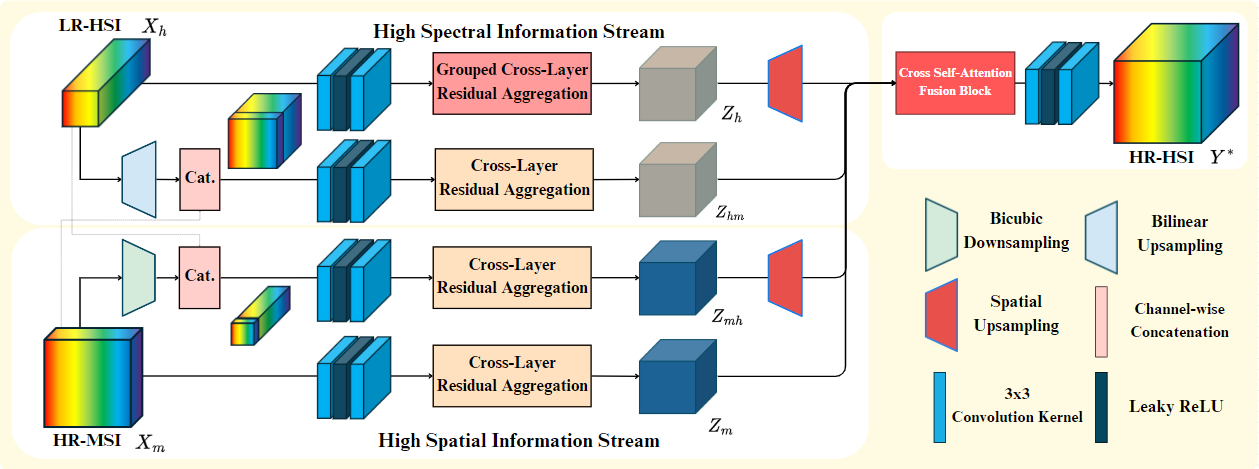}
\caption{{The proposed network architecture for HSI/MSI fusion based on the proposed Cross-Layer Residual Aggregation (CLRA) unit and Cross-Self-Attention (CSA) Fusion module. With the proposed Dual-two-Streamed (DTS) network, our network can judiciously learn the spatial-spectral representation across different branches. Afterwards, CSA enables network to adaptively fuse these representation, thereby yielding great results. By the proposed Knowledge Distillation (KD) manner, the network not only keep great performance, but reduce the model-size to fit real-world scenarios.}}
\label{fig:net}
\end{figure*}

Figure \ref{fig:net} gives the overview of the proposed lightweight deep network model for real-time HSI/MSI fusion tasks. First, a complex network, coupled with the proposed DTS (Dual Two-Streamed) network, is used as the teacher network. Then, a reduced version of our teacher network, with a reduced number of channels of each layer, is treated as the student network. In our network design, the proposed CSA (Cross Self-Attention) fusion module is essential for judiciously fusing the high-fidelity spatial and spectral quasi-fused (or initially fused) results generated from the proposed DTS backbone network. This design, incorporating different sampling rates in the spatial and spectral domain of HSI and MSI, respectively, could effectively capture the high-resolution spectral and spatial features simultaneously, thereby improving the performance without increasing the network complexity. A standard KD (knowledge distillation) loss is then applied to train the teacher and student networks simultaneously. Finally, the student network could fuse the HSI and MSI in real-time. The technical details will be revealed in the following subsections.

\subsection{Network Architecture}
To design a lightweight network model for efficient LR-HSI/HR-MSI fusion, we aim at leveraging knowledge distillation to reduce the network complexity. However, the KD often requires the network architectures of the teacher and student to be identical, so the native network architecture should be efficient and effective to have enough space to be pruned. Inspired by conventional ensemble learning in the machine learning field, it is possible to improve the classification performance by leveraging multiple independent classifiers together, where each classifier could be simple enough. Similarly, our DTS inherits the advantages of the idea for ensemble learning but rather just simply aggregates the spectral and spatial information from the input LR-HSI/HR-MSI. Specifically, we fully exploit spectral and spatial information from the inputs to design the four different sub-networks for information aggregation to improve the fusion performance with lower computational and space complexity. 

This Section illustrates the architecture design of our DTS network, consisting of spatial- and spectral-aware networks (SpaNet and SpeNet) for LR-HSI and HR-MSI, respectively, as shown in Figure \ref{fig:net}. To effectively refer the LR-HSI and HR-MSI information jointly, the proposed SpaNet and SpeNet not only retrieve the respective feature representations from LR-HSI and HR-MSI, but also extract the joint features of LR-HSI and HR-MSI simultaneously by internal feature concatenation, as shown in the left part of Figure \ref{fig:net}. In this way, we could easily integrate the spatial and spectral feature representation without increasing the model complexity. Then, a novel CSA fusion module is proposed to judiciously aggregate the spatial and spectral feature representations to obtain the final HR-HSI. The details can be found as follows.

\subsubsection{Proposed Dual Two-Streamed Network}
This subsection explicates the network architecture design, starting from the HR-HSI denoted as \(\mathbf{Y}\). The observable LR-HSI is modeled as \(\mathbf{X}_h = \mathbf{Y}\mathbf{B}\), where \(\mathbf{B}\) is the blurring matrix reducing pixel count. The observable HR-MSI is represented as \(\mathbf{X}_m = \mathbf{D}\mathbf{Y}\), with \(\mathbf{D}\) being the downsampling matrix that diminishes the number of spectral bands. 

Let the LR-HSI and HR-MSI be $\bX_{h} \in \mathbb{R}^{h_h\times w_h\times b}$ and $\bX_m \in \mathbb{R}^{h\times w\times b_m}$, the reconstructed HR-HSI denotes $\bY^* \in \mathbb{R}^{w\times h\times b}$ by the proposed DTS network using 
\begin{equation}
    \bY^*=f_{\text{DTS}}(\bX_h, \bX_m; \bW_{\text{DTS}}), 
\end{equation}
where $\bW_{\text{DTS}}$ is the weights of the proposed DTS network. As mentioned previously, we respectively sample the spatial and spectral features from $\bX_{h}$ and $\bX_{m}$ to have better fusion results. We start with spatial feature extraction for HR-MSI and LR-HSI. First, the high-spectral-resolution feature could be obtained as follows:
\begin{equation}
    \bZ_m = f_\text{CLRA}(\bX_m),
\end{equation}
where $f_{\text{CLRA}}$ denotes the proposed Cross-Layer Residual Aggregation (CLRA) module, and we will discuss CLRA later. As we were required to learn the fine detail features from LR-HSI and HR-MSI, we could upsample the LR-HSI $\bX_{h}$ to obtain the $\bX_{h}^{u} \in \mathbb{R}^{h\times w\times b} = f_{\text{up}}(\bX_{h})$, where $ f_{\text{up}}$ denotes the Bicubic interpolation function. 

Then, the fused spatial draft $\bZ_{hm}$ could be obtained by
\begin{equation}
    \bZ_{hm} = f_{\text{CLRA}} (\text{Cat}.(\bX_{h}^{u}, \bX_m)),
\end{equation}
where $\text{Cat.}$ is the channel-wise feature map concatenation operation. In this way, we smartly aggregate the HR-MSI and LR-HSI simultaneously in $\bZ_{hm}$, thereby improving the spatial quality of the reconstructed HR-HSI $\bZ^*$.

Meanwhile, the high-spectral-resolution information could also be obtained in a similar manner. Specifically, we have LR-HSI $\bX_h$, a rich spectral information data, that could be used to restore the spectrums for the reconstructed HR-HSI $\bY^*$. First, let the feature representation of $\bX_h$ be $\bZ_h$, we could simply obtain this feature representation by

\begin{equation}
    {{\bZ_h = f_{\text{CLRA}}(\bX_h; g_c),}}
\end{equation}
where $g_c$ indicates the group number for the grouped convolution operator in the proposed CLRA (will be discussed later). It is somewhat reasonable that the spectral redundancy is relatively high in the HSI, especially in the successive spectrums. Therefore, it is natural that the grouped convolution could be used to reduce complexity and maintain performance. 
While the spectral information could still be extracted from HR-MSI $\bX_m$, we followed a similar protocol to jointly retrieve the joint feature representation from both LR-HSI and HR-MSI by

\begin{equation}
    \bZ_{mh} = f_{\text{CLRA}} (\text{Cat}.(\bX_{m}^{d}, \bX_h)),
\end{equation}
where $\bX_m^d$ denotes the spatially dowsampled HR-MSI $\bX_m$ by Bicubic interpolation function $f_{\text{down}}$. In this way, the high quality spectrum information should be able to reconstruct by merging $f_{\text{up}}(\bZ_{mh})$ and $f_{\text{up}}(\bZ_{m})$.
It is easy to obtain the fused HR-HSI by simple modalities ensemble by
\begin{equation}
    \bY^* = f_{\text{up}}(\bZ_{mh}) + f_{\text{up}}(\bZ_{h}) + \bZ_m + \bZ_{hm}.
\end{equation}

However, different modalities, $\bZ_{mh}$, $\bZ_{m}$, $\bZ_m$, and $\bZ_{hm}$ might exist conflict in spatial or spectral features so that the performance could be suppressed. Moreover, retaining the high quality of the reconstructed HR-HSI in noised HR-MSI or LR-HSI is essential and desired. If HR-MSI or LR-HSI has been perturbed by random noise during transmission or sensor noise, the performance of the reconstructed HR-HSI could be degraded significantly. Considering that the noise $\bN \sim N(\mu, \sigma)$ with zeros mean $\mu=0$ and a standard deviation $\sigma$, the noised LR-HSI $\bX_h$ could be $\bX'_h=\bX_h+\bN$. In this case, the feature representations, $\bZ_m$, $\bZ_{mh}$, and $\bZ_{hm}$, could be also noised propagated. So, the fused HR-HSI could be

\begin{equation}
\begin{aligned}
    \bY^* &= f_{\text{up}}(\bZ'_{mh}) + \bZ'_{hm} + \bZ'_m + f_{\text{up}}(\bZ_{h})\\
    & + [f_{\text{up}}(\bZ_{mh}) + f_{\text{up}}(\bN_{mh})] + [f_{\text{up}}(\bZ_{hm}) + f_{\text{up}}(\bN_{hm})] \\
    & + (\bZ_{m} + \bN_{m}) + \bZ_h
\end{aligned}
\end{equation}
where $f_{\text{up}}(\bN_{hm})$, $f_{\text{up}}(\bN_{mh})$, and $f_{\text{up}}(\bN_{m})$ represent the feature representation of noise pattern $\bN$. While the additive noise during the training phase might enhance the robustness of the reconstructed HR-HSI for the proposed DTS network, the equal weights shared with four modalities still lead to restricted performance. A smart way to adaptively fuse these modalities would be to use dynamic weights instead of equal weights for better performance and robustness, i.e., the proposed CSA Fusion Module. 

\begin{figure}[htbp]
\centering
\includegraphics[width=0.7\linewidth]{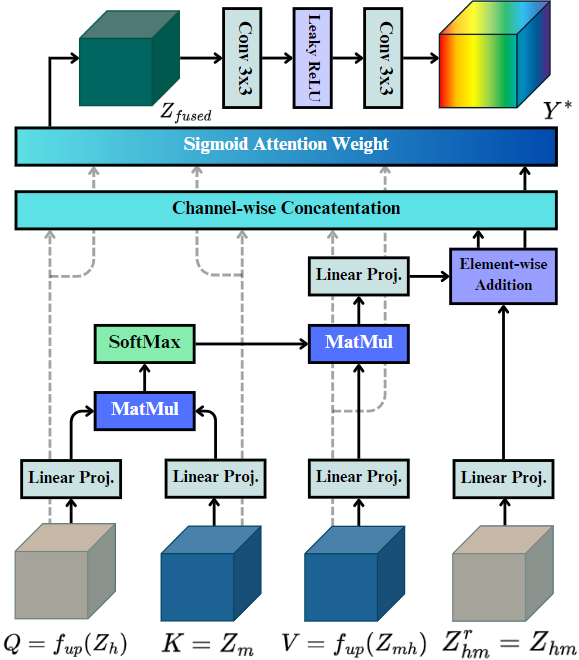}
\caption{The proposed Cross self-attention (CSA) fusion module. The blue cube contains high-spatial information, and the other two contain relatively rich spectral information. The proposed attention module smartly considers the weight of different branches and fuses these representations together.}
\label{fig:csa}
\end{figure}

\subsubsection{Proposed Cross-Self-Attention Fusion Module}
To learn the adaptive weight across different modalities of features, we propose a novel attention module to fuse the feature representations of LR-HSI and HR-MSI judiciously, as shown in Figure \ref{fig:csa}. To reduce the computational complexity of the high-dimensional feature representation like $\bZ_m$ and $\bZ_{hm}$, a simple bottleneck layer is used to project the feature maps to the lower-dimensional ones. 
Suppose that the reduced features of $f_{\text{up}}(\bZ_h)$, $\bZ_m$,  $f_{\text{up}}(\bZ_{mh})$, and $\bZ_{hm}$ denoted by $\bQ$, $\bK$, $\bV$, and $\bZ_{hm}^r$, the projection is defined as follows:

\begin{equation}
\begin{aligned}
\bQ_i &= \text{Proj.}(\bQ) \\
\bK_i &= \text{Proj.}(\bK) \\
\bV_i &= \text{Proj.}(\bV) \\
\bZ_{hm}^r &= \text{Proj.}(\bZ_{hm}) \\
\end{aligned}
\end{equation}
where $\text{Proj.}(\cdot)$ consists of multiple stages to project the input feature into lower dimensional space. First, the $1\times 1$ convolution is used to project the $\bX \in \mathbb{R}^{b\times c \times h \times w}$ into $\bX' \in \mathbb{R}^{b\times r\times h\times w}$, where $r$ is the reduced number of dimension. To enable the multi-head attention in CSA, we reshape the $\bX'$ into a feature vector sized of $b\times h_a \times r_o$, where $r_o$ is determined by $r/h_a$, and $h_a$ denotes the number of multi-heads in the attentions. Now, we could perform the cross-attention by 

\begin{equation}
    \begin{aligned}
\bA_i &= \text{softmax}\left(\frac{\bQ_i \cdot \bK_i^T}{\sqrt{r}}\right), \\
\bO &= \text{Cat.}(\bA_0 \cdot \bV_0, \bA_1 \cdot \bV_1, ..., \bA_{h_a} \cdot \bV_{h_a}) , \\
\bO &= \text{Proj.}_O(\bO) + \bZ_{hm}^r, \\
\bC &= \text{Cat.}(\bQ, \bK, \bV, \bO), \\
\bW &= \text{Sigmoid}(\text{Proj.}_C(\bC)), \\
    \end{aligned}
\end{equation}
where $\text{Proj.}_O$ aims to project the concatenated multi-head attentions into the same dimension with $\bZ_{hm}^r \in \mathbb{R}^{b \times c \times h \times w}$, $\text{Proj.}_C$ projects the concatenated cross-attentions to adaptive weights $\bW$, i.e., $b\times 4\times h\times w$. In this way, we could judiciously fuse the different modalities, and, even under noised inputs, the proposed CSA still remains strong due to its adaptivity, as follows:
\begin{equation}
    \bZ_{\text{fused}} = \bW_1 \cdot \bQ + \bW_2 \cdot \bK + \bW_3 \cdot \bV + \bW_4 \cdot \bO,
\end{equation}
where $\bW_i$ indicates $i$-th channel of $\bW$. 
Finally, the reconstructed HR-HSI is obtained via a simple convolution layer by
$\bY^* = \text{Conv}_{\text{HR}} (\bZ_{\text{fused}})$.

\subsubsection{Cross-Layer Residual Aggregation Module}
\begin{figure}[htbp]
\centering
\includegraphics[width=1\linewidth]{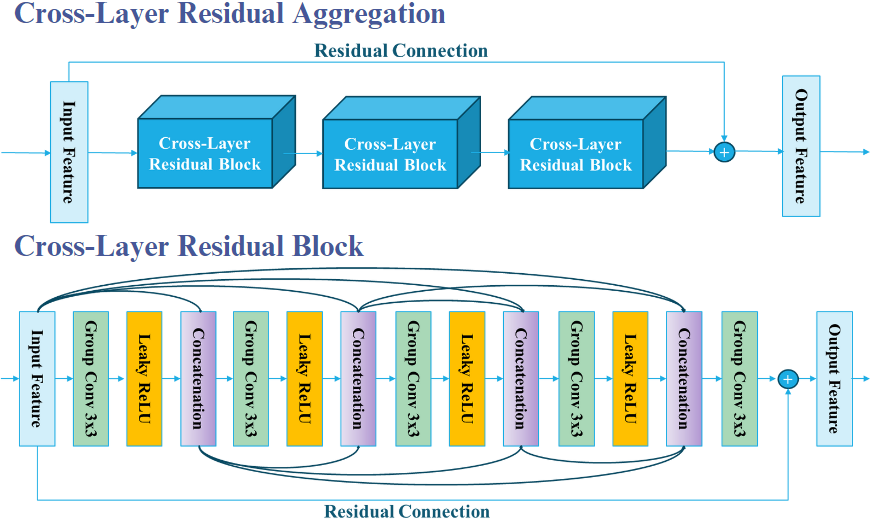}
\caption{The overview of proposed CLRA. Each CLRA contains three CLRB and residual connection. As for CLRB, it is stacked by several convolution operators, such as LeakyReLU, dense, and residual connections.}
\label{fig:clrb}
\end{figure}

Designing an effective and efficient block to capture the spatial and spectral features of HSI is essential. This subsection aims to draw a basic block design of our CLRA, as shown in Figure \ref{fig:clrb}. In the proposed CLRA module, inheriting the advantages from the block designed in DCSN \cite{36}, the residual connection and densely connected feature concatenation are also adopted to make the larger receptive field in the single CLR block, as shown in the bottom part in Figure \ref{fig:clrb}. By aggregating three CLR blocks with a residual connection between the input and output features, we could form the basic CLRA module, as shown in the top part in Figure \ref{fig:clrb}. Consider that the high-spectral-resolution input, i.e., $\bX_h$, has high redundancy between the successive bands, the grouped convolution operation is adopted in our CLRA for $\bX_h$ with group number $g_c$, while other inputs remain to adopt normal convolutional operation. Note that the merged high-spectral-resolution input, i.e., $\bX_{hm}$ and $\bX_{mh}$, also adopts the standard convolution since there might exist the useful information between $\bX_h$ and $\bX_m$. 

On the other hand, to have a lower latency in the inference phase, relatively shallow networks are constructed for the four input branches by stacking our CLRA by 6, 6, 4, and 4 times for extracting the feature $\bZ_{hm}, \bZ_{mh}, \bZ_{h}, \text{and} \bZ_m$ in our teacher model, respectively. Conversely, the student model stacks CLRA by 1, 4, 4, and 1 times to reduce its computational complexity.



\subsection{Joint Training via Knowledge Distillation}


Traditional knowledge distillation (KD) techniques often employ a feature map-based loss, where the student network is trained to mimic the intermediate feature representations of the teacher network. This method, while effective in some scenarios, imposes a stringent requirement on the student network to replicate the feature maps exactly as the teacher's. Such a constraint can limit the learning capacity of the student network, particularly when the student's architecture is much lighter, and may lead to difficulties in convergence due to the complex nature of the feature spaces involved.

The feature map-based KD loss assumes that a direct correspondence between the teacher and student feature maps is necessary for knowledge transfer. However, this can be overly restrictive, as the student network might benefit from developing its unique feature representations that are more suited to its capacity, yet still retain the essential characteristics learned by the teacher. The forced alignment of feature maps can, therefore, be counterproductive, leading to a challenging training process and potentially suboptimal student performance.

To address these issues, the feature-map KD loss should be placed in the relatively rare layers instead of each layer to allow the student network to learn its unique feature representations in most layers, thereby improving the performance of the student network. Specifically, Sigmoid cross-entropy loss is used to approximate the feature map distributions of student and teacher networks, as follows:

\begin{equation}
\begin{aligned}
    \ell_{\text{KD}} &= -(f_{\text{s}}(\bZ_{\text{fused}}^s)\log(f_{\text{s}}(\bZ_{\text{fused}}^t) \\
    &+ (1 - f_{\text{s}}(\bZ_{\text{fused}}^s)\log(1 - f_{\text{s}}(\bZ_{\text{fused}}^t))
\end{aligned}
\end{equation}
where $f_{\text{s}}$ denotes the Sigmoid activation function, and ${\bZ_{\text{fused}}^s}$ and ${\bZ_{\text{fused}}^t}$ represent the fused feature maps from student and teacher networks. To leverage the good quality of the reconstructed HR-HSI, the reconstruction-relative loss functions should be involved to enhance the spectral and spatial quality. Traditionally, the $\ell$-1 norm distance metric aims to enhance the data fidelity, while the energy of each band in an HSI may vary significantly, leading to the that the traditional $\ell_{\text{L1}}$ distance could pay more attention to the bands whose energy is relatively large. However, the spectrum feature of HSIs is essential for different tasks since each band has its purposes. Therefore, we propose a Band-Energy-Balance-Aware (BEBA) loss $\ell_{\text{BEBA}}$ to judiciously facilitate the problem above, thereby improving the spectrum quality of the reconstructed HSI $\bY^*$.  

\begin{equation}
\begin{aligned}
\ell_{\text{BEBA}} = \frac{f_\text{m}\left(\alpha \bD / \beta + f_{\text{ReLU}}(\bD - \beta) - \alpha\beta \right)}{f_{\text{m}}(\bY^2 + \epsilon)},
\end{aligned}
\end{equation}
where $\alpha$ and $\beta$ are regularization parameters, $\bD$ denotes the squared absolute difference between the prediction and target $|\bY^* - \bY|^2$, $\epsilon$ is a small positive constant, and $f_{\text{m}}$ denotes the mean operator over spatial axis. Specifically, $\alpha=0.5$ and $\beta=1$ are chosen in our experiments. In this way, the $f_{\text{m}}(\bY^2 + \epsilon)$ captures the energy of each band, thereby dynamically adjusting the weights of each band according to its energy. 

The parameters $\alpha$ and $\beta$ play crucial roles in balancing the sensitivity of the loss function towards small and large prediction errors. The term $\alpha$ primarily scales the mean squared error, enhancing the function's reactivity to smaller deviations between the predicted $\bY^*$ and the ground truth HR-HSI $\bY$. This scaling is particularly significant when dealing with data that possess subtle variations, as it amplifies the importance of minor discrepancies.

The parameter $\beta$, on the other hand, serves as a thresholding value that delineates the boundary between small and large errors. When the squared difference $\bD$ is less than $\beta$, the ReLU term $f_{\text{ReLU}}(\bD - \beta)$ becomes zero, and the loss function primarily operates in a quadratic regime dominated by $\alpha \bD / \beta$. This regime is sensitive to smaller errors, thus ensuring precision in the predictions. Conversely, for larger errors where $\bD$ exceeds $\beta$, the loss function transitions into a linear regime, mitigating the potential issues of gradient explosion typically associated with large errors in quadratic loss functions. This linear portion of the loss function is given by $f_{\text{ReLU}}(\bD - \beta) - \alpha\beta$, which acts as a safeguard against the disproportionate penalization of large errors, enhancing the robustness of the model against outliers and noise.

To enhance the spectral quality of the reconstructed HR-HSI further, Spectral Angle Mapper (SAM) loss $\ell_{\text{SAM}}$ is also proposed to guide our teacher and student networks, as follows:
\begin{equation}
\ell_{\text{SAM}}=
1 - \frac{1}{HW}
\sum_{n=1}^{HW}
\left(
\frac{(\bm {Y}_n)^T{\bm{Y}^*_n}}{|\bm {Y}_n|_2\cdot|\bm {Y}^*_n|_2 + \epsilon}
\right),
\end{equation}
where $\bY_n$ denotes the $n$-th spectral vector, and we calculate the negative cosine similarity between the reconstructed HR-HSI $\bY^*$ and the ground truth HR-HSI $\bY$ as the SAM loss. This measure effectively captures the angular difference between the spectral signatures in the hyperspectral data, making it a robust metric for assessing the spectral fidelity of the predicted image in comparison to the ground truth. The cosine similarity is computed as the dot product of the vectors, normalized by the product of their magnitudes, ensuring that the loss function focuses solely on the angular difference, independent of the magnitude of the spectral signatures.

Finally, the standard reconstruction loss, i.e., $\ell$-1 norm loss $\ell_{\text{L1}}$, is used to ensure the high quality of the reconstructed HSI. Thus, the total loss of the teacher network would be
\begin{equation}
\begin{aligned}
    \ell_{\text{t}} &= \ell_{\text{L1}} (\bY^t, \bY) + \lambda_1 \ell_{\text{BEBA}}(\bY^t, \bY) \\
    &+ \lambda_2 \ell_{\text{SAM}}(\bY^t, \bY),
\end{aligned}
\end{equation}
where the superposition of $\bY^t$ denotes the reconstructed HR-HSI by our teacher network and $\lambda$ is the parameters to control the importance between the spatial fidelity and spectral quality terms. Likewise, the total loss of the student network is defined by the reconstruction loss and KD loss, as follows:

\begin{equation}
\begin{aligned}
    \ell_{\text{s}} &= \ell_{\text{L1}} (\bY^s, \bY) + \lambda_1 \ell_{\text{BEBA}}(\bY^s, \bY) \\ 
    &+ \lambda_2 \ell_{\text{SAM}}(\bY^s, \bY) + \lambda_3 \ell_{\text{KD}}((\bY^s, \bY)) \\
    &+ \lambda_4 \ell_{\text{L1}} (\bY^s, \bY^t),
\end{aligned}
\end{equation}
where $\ell_{\text{L1}} (\bY^s, \bY^t)$ aims to relax the constraint of $\ell_{\text{L1}} (\bY^s, \bY)$ since the outcome of lightweight student network might hard to approximate to the ground truth accurately. All the balance parameters, $\lambda_1$,$\lambda_2$,$\lambda_3$, and $\lambda_4$, are set to $0.1$ respectively. 


\section{Experimental Results}\label{sec:expALL}

\subsection{Experiment Settings}

\subsubsection{Dataset Preparation and Synthesis of LR-HSI and HR-MSI}
The dataset used for performance evaluation in this study was acquired by the Airborne Visible/Infrared Imaging Spectrometer (AVIRIS) sensor \cite{35}. The collected dataset includes various natural landscapes from the US and Canada, such as cities, mountains, lakes, fields, and plants, captured between 2006 and 2011. The original HSI images were partitioned into non-overlapping sub-images of size $259 \times 259$ pixels, with 224 spectral bands covering a wavelength range from 400 to 2500 nm. As suggested in \cite{36}, low-quality bands (1-10, 104-116, 152-170, and 215-224) were removed, resulting in HSI images with 172 spectral bands.

To simulate the image fusion experiments, Wald's protocol \cite{37} was employed. The HR-HSI $\bY$ of size $256 \times 256$ pixels was cropped from the top-left corner of each HSI sub-image. Two different downsampling matrices were used to synthesize multispectral images (MSI) $\bY_{m4}\in\mathbb{R}^{256\times 256\times 4}$ and $\bY_{m6}\in\mathbb{R}^{256\times 256\times 6}$ with four and six spectral bands, respectively. The downsampling matrix $\bD_4\in\mathbb{R}^{4\times 172}$ approximately corresponds to Landsat TM bands 1-4 (covering 450-520, 520-600, 630-690, and 770-900 nm), while the downsampling matrix $\bD_6\in\mathbb{R}^{6\times 172}$ roughly corresponds to Landsat TM bands 1-5 and 7 (covering 450-520, 520-600, 630-690, 770-900, 1550-1750, and 2090-2350 nm).
A Gaussian point spread function with a variance of $\sigma=3$ and a blurring factor of $b_r=4$ was used to generate a spatially degenerated matrix $\bB\in\mathbb{R}^{L^2\times {L_l}^2}$ for synthesizing LR HSIs. The LR-HSI $\bY_{h1} \in \mathbb{R}^{64\times 64\times 172}$ was obtained by applying the spatially degenerated matrix $\bB$ to the HR-HSI $\bY$. 

The collected dataset consisted of 2,078 HR-HSI images, which were randomly partitioned into training, validation, and testing sets for performance evaluation. The training set contained 1,678 images, while the validation and testing sets contained 200 images for each. The spatial and spectral resolutions of the HR-MSI and LR-HSI were $256\times 256 \times M_m$ and $64\times 64\times 172$, respectively, where $M_m$ is either 4 or 6 in our experiments.

\subsubsection{Implementation Details.} The experimental platform utilized in this study comprised an Intel\textregistered Xeon\textregistered Gold 61 CPU, 90GB of system memory, and an NVIDIA Tesla V100 GPU with 32GB of memory. The proposed method was implemented using the PyTorch deep learning framework. The batch size was set to 4, and the number of training epochs was fixed to 600 for all experiments involving the proposed method. For the peer methods, the number of training epochs was set according to their default values as specified in their respectively original publications. The online distillation strategy employed in the proposed framework facilitated simultaneous updates of the teacher and student networks during the training process. The Adam optimizer \cite{38} was used for training, with an initial learning rate of 0.0001. The learning rate was adjusted during the training process using the Cosine Annealing learning decay scheduler. The weights of the penalty terms in the loss function, denoted as $\lambda_s$ and $\lambda_t$, were both set to 0.1. Standard data augmentation, including random cropping and rotation, is adopted in this paper for all of the evaluated methods.

\subsubsection{Quantitative Metrics} For better comprehensive evaluation, we adopt the three commonly-used quantitative metrics: 

\begin{enumerate}

    \item Peak signal-to-noise ratio (PSNR in dB) is defined as
    \begin{equation*}
    \text{PSNR}=\frac{1}{M}\sum_{m=1}^{M}\text{PSNR}_m,
	\end{equation*}
	where $\text{PSNR}_m$ measures the spatial quality of a single band, and ${m}$ represents the ${m}$-th band, defined by:
	\begin{equation*}
	\text{PSNR}_m \!=\!
	10\log_{10}\left(\frac{\max\{\bm {y}_{mn}^2\mid n\in\mathcal{I}_L\}}{\frac{1}{L}\|{\bm {Y}}^{(m)}-{{\bm {Y}}}^{*(m)}\|_2^2}\right),
	\end{equation*}
	where $\bm {z}_{mn}$ denotes the $n$th entry in the vector $\bm {Y}^{(m)}$, and $\mathcal{I}_L\triangleq\{1,\dots,L\}$.
	A higher $\text{PSNR}$ value indicates a better spatial quality of the fused image $\bm {Y}^*$;
    \item Spectral angle mapper (SAM) is defined as
	\begin{equation*} 
	\text{SAM}=
	\frac{1}{L}
	\sum_{n=1}^{L}
	\text{arccos}\left(
	\frac{(\bm {y}[n])^T{\bm^*{y}}[n]}{\|\bm {y}[n]\|_2\cdot\|\bm {y}^*[n]\|_2}
	\right),
	\end{equation*}
	where $\bm{y}[n]$ denotes the $n$th column of $\bm {Y}$.
	The lower the absolute value of SAM is, the greater the spectral restoration performance of $\bm {Y}^*$ is; and

	\item Root mean squared error (RMSE) is defined as
	\begin{equation*} 
	\text{RMSE}=
	\sqrt{
	\frac{1}{M}
	\sum_{m=1}^{M}
	\text{RMSE}_{m}^2},
	\end{equation*}
	where  
	\begin{equation*} 
	\text{RMSE}_{m}=
	\frac{1}{\sqrt{L}}
	{\|\bm {Y}^{(m)} - \bm {Y}^{*(m)}\|_2},
	\end{equation*}
	
	The smaller the RMSE value is, the better the global quality of the fused image $\bm {Y}^*$ is.

\end{enumerate}

\subsection{Performance Evaluation}

\begin{figure*}[htbp]
\centering
\includegraphics[width=0.8\linewidth]{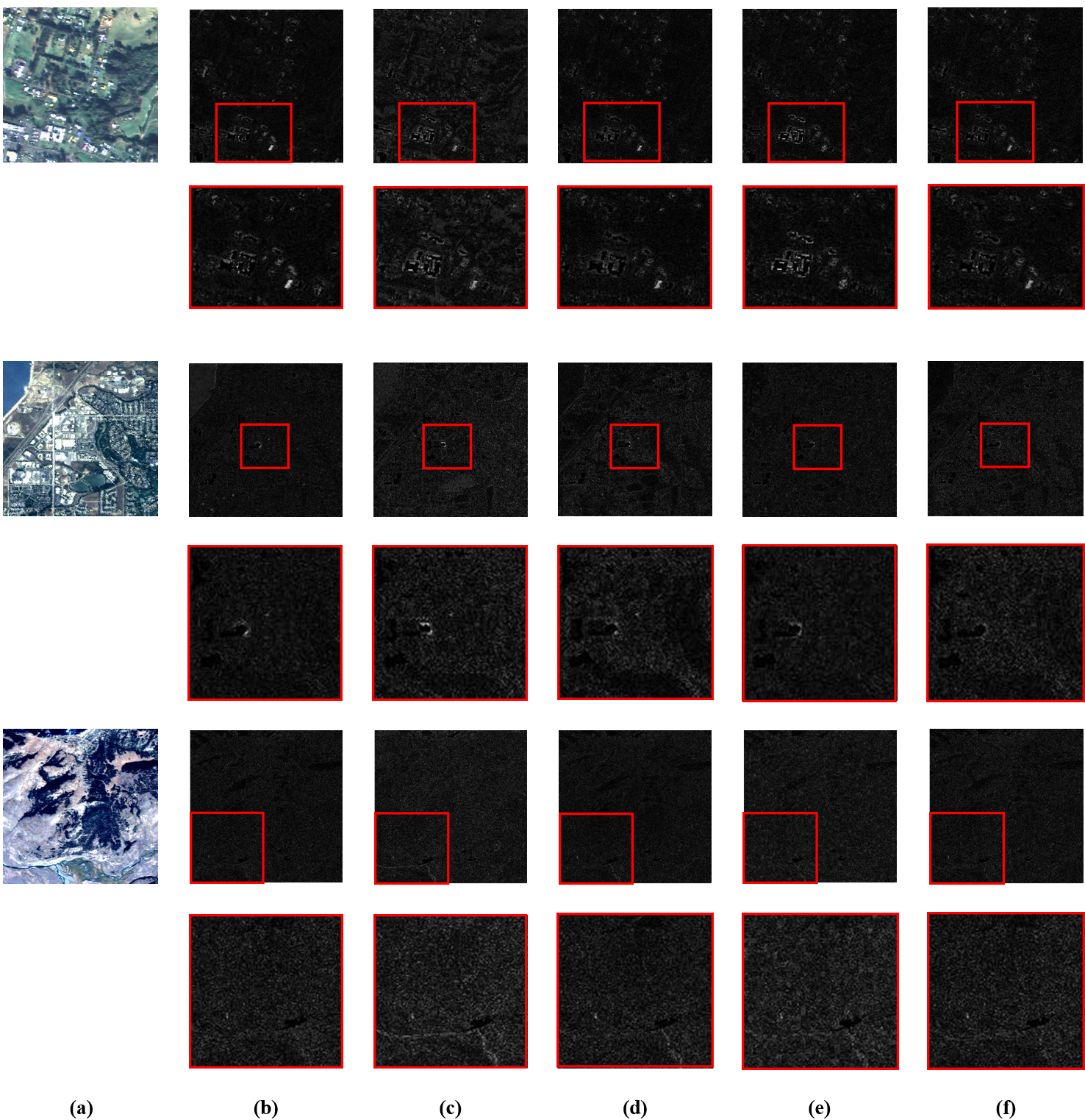}
\caption{Hyperspectral and Multispectral fusion images at AVIRIS dataset. The upper row is the fused RGB image, and the lower row is the residual image subtracted from Ground Truth : (a) the Ground Truth image ; (b) the Proposed method ; (c) PZRes-Net \cite{24} ; (d) MSSJFL \cite{23} ; (e) Dual-UNet \cite{25} ; (f) DHIF-Net \cite{26}.}
\label{fig:exp1}
\end{figure*}

\begin{figure*}
\centering
\includegraphics[width=1\linewidth]{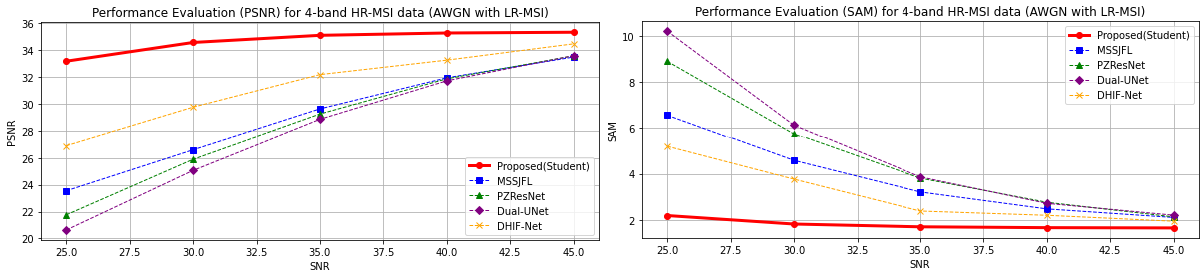}
\caption{Robustness comparison among the proposed method and other peer methods under different SNR values in LR-HSI for 4-band HR-MSI.}
\label{fig:robust4}
\end{figure*}

\begin{figure*}
\centering
\includegraphics[width=1\linewidth]{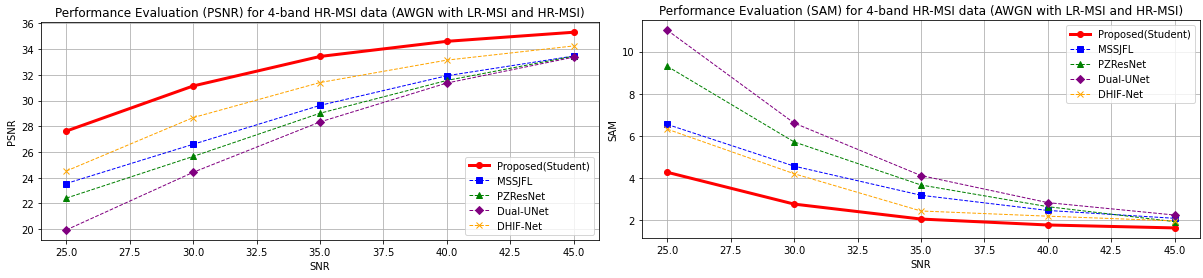}
\caption{Robustness comparison among the proposed method and other peer methods under different SNR values in both LR-HSI and 4-band HR-MSI.}
\label{fig:robust4a}
\end{figure*}

\begin{figure*}
\centering
\includegraphics[width=1\linewidth]{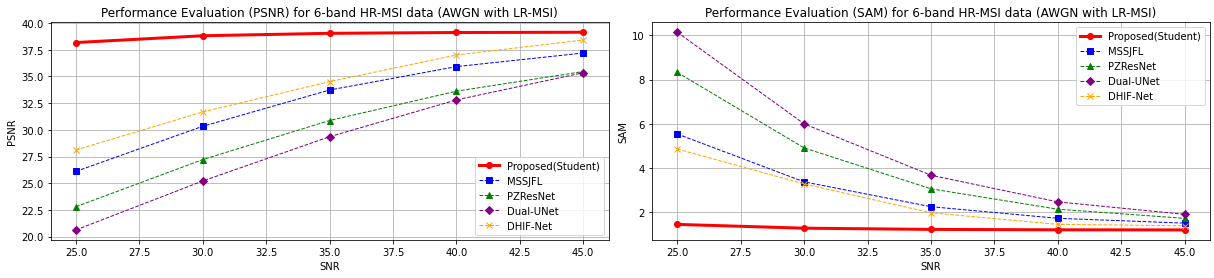}
\caption{Robustness comparison among the proposed method and other peer methods under different SNR values in LR-HSI for 6-band HR-MSI.}
\label{fig:robust6}
\end{figure*}

\begin{figure*}
\centering
\includegraphics[width=1\linewidth]{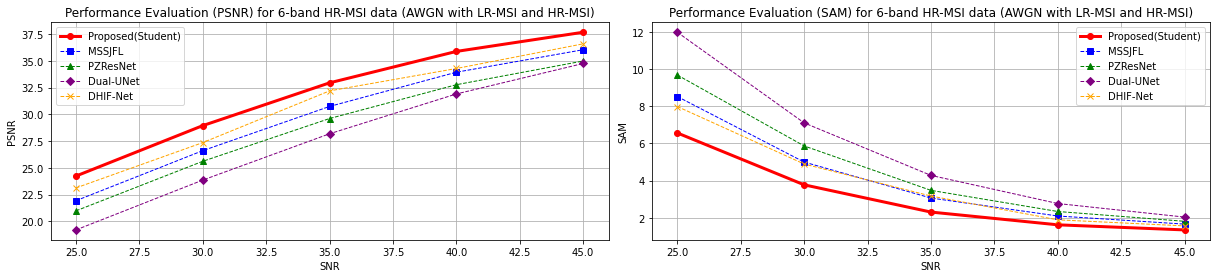}
\caption{Robustness comparison among the proposed method and other peer methods under different SNR values in both LR-HSI and 6-band HR-MSI.}
\label{fig:robust6a}
\end{figure*}

\begin{table*}[htbp]
\centering
\caption{Performance evaluation and complexity comparison of the proposed method and other fusion models in terms of several metrics. Note that the methods marked with an asterisk (*) are unsupervised approaches. For the complexity parts, M and G indicate $10^6$ and $10^9$, respectively. \textbf{L} denotes the large version. EXT represents the extended training scenario, where we reduced the learning rate to $5e-5$ and trained for an additional 40 epochs.}
\label{tab:performance}
\scalebox{0.98}{
\begin{tabular}{l|c|ccc|ccc|cccc}
{} &  & \multicolumn{3}{l}{4 Bands LR-HSI} & \multicolumn{3}{l}{6 Bands LR-HSI} & \multicolumn{3}{l}{4 Bands LR-HSI}\\\hline
Method &  Venue & PSNR↑ &  SAM↓ &  RMSE↓  &              PSNR↑ &  SAM↓ &  RMSE↓ & Params & FLOPs & Run-time & Memory\\\hline
PZRes-Net\cite{24} & TIP 2021 & 34.963 &  1.934 &  35.498 & 37.427 &  1.478 &  28.234 & 40.15M & 5262G & 0.0141s & 11059MB \\
MSSJFL\cite{23}    & HPCC 2021 &  34.966 &  1.792 &  33.636 & 38.006 &  1.390 &  26.893 & 16.33M & 175.56G & 0.0128s & \best{1349M} \\
Dual-UNet \cite{25} & TGRS 2021 & 35.423 &  1.892 &  33.183 &  38.453 &  1.548 &  26.148 & \best{2.97M} & \best{88.65G} & \secondBest{0.0127s} & 2152M \\
DHIF-Net  \cite{26} & TCI 2022 & 34.458 &  1.829 &  34.769 & 39.146 &  1.239 &  25.309 & 57.04M & 13795G & 6.005s & 29381M\\
*CUCaNet   \cite{43} & ECCV 2020  &  28.848 &  4.140 &  71.710 & 35.509 &  2.205 &  38.973 & 3.0M & 40.0G & 2070.01s & -\\
*USDN   \cite{44}    & CVPR 2018 &  30.069 &  3.688 &  93.408 & 35.208 &  2.650 &  53.987 & 0.006M & 1.0G & 28.83s & - \\
*U2MDN  \cite{45}  & TGRS 2021 &  30.127 &  3.235 &  59.071 & 33.356 &  2.243 &  41.528 & 0.01M & 4.0G &
 547.28s & -\\\hline
\textbf{Proposed-Teacher}&   -    &  \best{35.967} &  \best{1.527} &  \best{30.928} &  \best{40.046} &  \best{1.095} &  \best{23.785} & 26.8M & 941.77G & 0.0134s & 8733M\\
\textbf{Proposed-Student}&   -  &  \secondBest{35.544} &  \secondBest{1.643} &  \secondBest{32.308} &  \secondBest{39.153} &  \secondBest{1.205} &  \secondBest{25.080} & \secondBest{7.44M} & \secondBest{144.77G} & \best{0.0121s} & \secondBest{1653M}\\\hline
\textbf{Proposed-Teacher-L}&   -    &  \best{36.098} &  \best{1.503} &  \best{30.577} &  \best{40.048} &  \best{1.092} &  \best{23.733} & 37.19M & 1303.3G & 0.1117s & 12110M\\
\textbf{Proposed-Student-L}&   -  &  \secondBest{35.548} &  \secondBest{1.588} &  \secondBest{31.561} &  \secondBest{39.784} &  \secondBest{1.119} &  \secondBest{23.956} & 11.34M & 399.9G & 0.0292s & 4054M\\\hline
\textbf{Proposed-Teacher-L-Ext}&   -    &  \best{36.076} &  \best{1.508} &  \best{30.589} &  \best{40.043} &  \best{1.098} &  \best{23.754} & 37.19M & 1303.3G & 0.1117s & 12110M\\
\textbf{Proposed-Student-L-Ext}&   -  &  \secondBest{35.954} &  \secondBest{1.528} &  \secondBest{30.801} &  \secondBest{39.801} &  \secondBest{1.115} &  \secondBest{23.844} & 11.34M & 399.9G & 0.0292s & 4054M
\end{tabular}}
\end{table*}

\begin{table*}[]
\centering
\caption{Robustness comparison among several state-of-the-art methods. This table illustrates the model performance for different noise inputs 4 bands LR-HSI and HR-MSI with the addition of AWGN noise. This table corresponds to Figure \ref{fig:robust4a}.}
\label{tab:robustness-eval}
\begin{tabular}{l|cccccc|c}
SNR Ratio & 25\% (Noisy) & 30\% & 35\% & 40\% & 45\% & 0\% (Clean) & Average \\ \hline
Method & PSNR / SAM & PSNR / SAM & PSNR / SAM & PSNR / SAM & PSNR / SAM & PSNR / SAM & PSNR / SAM \\ \hline
PZResNet \cite{24}& 22.417 / 9.3 & 25.658 / 5.72 & 29.017 / 3.681 & 31.566 / 2.654 & 33.454 / \secondBest{1.945} & 34.963 / 1.934 & 29.512 / 4.205 \\
MSSJFL \cite{23}& 23.553 / 6.549 & 26.603 / 4.573 & 29.627 / 3.195 & 31.932 / 2.475 & 33.464 / 2.103 & 34.966 / \secondBest{1.792} & 30.024 / 3.447 \\
Dual-UNet \cite{25}& 19.944 / 11.009 & 24.423 / 6.614 & 28.339 / 4.128 & 31.365 / 2.846 & 33.378 / 2.258 & \secondBest{35.423} / 1.892 & 28.812 / 4.791 \\
DHIF-Net \cite{26}& \secondBest{24.526 / 6.324} & \secondBest{28.677 / 4.214} & \secondBest{31.405 / 2.45} & \secondBest{33.148 / 2.204} & \secondBest{34.251} / 1.98 & 34.458 / 1.829 & \secondBest{31.077} / \secondBest{3.166} \\ \hline
\textbf{Proposed - Student} & \best{27.632 / 4.282} & \best{31.138 / 2.776} & \best{33.432 / 2.068} & \best{34.609 / 1.787} & \best{35.316 / 1.648} & \best{35.544 / 1.643} & \best{32.945 / 2.367}
\end{tabular}
\end{table*}

We evaluated our approach against seven state-of-the-art HSI/MSI fusion methods, including four supervised methods: PZRes-Net \cite{24}, MSSJFL \cite{23}, Dual-UNet \cite{25}, and DHIF-Net \cite{26}, and three unsupervised methods: CUCaNet \cite{43}, USDN \cite{44}, and U2MDN \cite{45}. The performance was objectively measured using the three metrics, including PSNR, SAM, and RMSE. Experiments were conducted with both 4 and 6 MSI bands. As more bands in the LR-HSI are available, richer spectral information can be exploited to potentially enhance the SR quality. The quantitative results are presented in Table \ref{tab:performance}. This study investigates two variants of the proposed model, denoted by the postfixes \textbf{L} and \textbf{L-Ext}. The \textbf{L} model is constructed by stacking additional blocks, as described in Section III.E, resulting in a larger model architecture. On the other hand, the \textbf{L-Ext} model is obtained by extending the training process of the \textbf{L} model with more epochs and a reduced learning rate of $5\times10^{-5}$.

The experimental results demonstrate that the proposed framework outperforms the compared state-of-the-art methods in terms of spectral reconstruction performance for both 4 and 6 bands of HR-MSI. Additionally, our method exhibits superior overall and pixel-level restoration capabilities compared to the state-of-the-art methods based on the obtained better performances in terms of the SAM, PSNR, and RMSE metrics, respectively. The outstanding performance of our method can be attributed to several factors. First, the proposed DTS network effectively integrates spatial-spectral feature representations, leading to the excellent performance of the teacher model. Second, the relatively shallow network architecture of the student model enables faster inference times while maintaining high-quality results, resulting in a higher performance-to-complexity ratio compared to previous methods. Third, the proposed response-based KD framework provides refined and strong guidance, facilitating the student network in learning nuanced representations with fewer parameters, thereby streamlining the architecture without compromising model performance. Furthermore, the proposed CSA fusion module and the distillation strategy enable our method to adaptively determine the optimal weights for HSI and MSI features, even in the presence of noise, resulting in improved robustness and stability. The effectiveness of the CSA and KD strategies will be further demonstrated in the subsequent experiments. On the other hand, the unsupervised learning approach, such as CUCaNet \cite{43}, USDN \cite{44}, and U2MDN \cite{45}, are hard to meet the requirements of real-time inference scenarios, even the relatively lower number of parameters and FLOPs (floating point operations).

Discerning differences in HSI images through visualization in the RGB color system is challenging. To better observe these differences, we compute residual images by subtracting each method's fused image $\bY^*$ from the corresponding ground truth $\bY$ and enhance the contrast through logarithmic mapping. The resulting residual images, which would be closer to black while being closer to the ground truths, are depicted in Figure \ref{fig:exp1}. These visualized results not only corroborate the superior quantitative performance of our method but also highlight the effectiveness of the proposed CSA and KD strategies in preserving fine image details and producing visually appealing false-color representations.



\subsection{Robustness Evaluation}
In real-world scenarios, the quality of LR-HSI and HR-MSI would suffer from lossy transmission or physical distortions, leading to the presence of heavy noise in the input data. Such noise would cause significant degradation and potentially catastrophic restoration results in the LR-HSI/HR-MSI fusion task. Therefore, it would be crucial to evaluate the robustness of the considered fusion methods in the noisy scenarios. 

To assess the robustness of each method, we introduced Additive White Gaussian Noise (AWGN) with varying Signal-to-Noise Ratios (SNR) ranging from 25 to 45. The AWGN noise and its impact on the HSI are formulated as:
\begin{equation}
\bN_{\text{awgn}} = \sqrt{\frac{\frac{1}{N}\sum_{i=1}^{N}{\bX_i^2}}{{10}^{\frac{SNR}{10}}}} 
\end{equation}
where $\bX_i$ represents the tensor-form input image and $N$ is the number of tensor elements. The noisy LR-HSI and HR-MSI can be obtained by $\bX_m=\bX_m+\bN_{\text{awgn}}$ and $\bX_h=\bX_h+\bN_{\text{awgn}}$, respectively.

As shown in Table \ref{tab:robustness-eval}, the proposed method achieves the best performance under noisy scenarios. We further considered two scenarios: (1) adding AWGN to the LR-HSI only, and (2) adding AWGN to both the LR-HSI and HR-MSI. The results for these scenarios are presented in Figures \ref{fig:robust4}, \ref{fig:robust4a}, \ref{fig:robust6}, and \ref{fig:robust6a}. As expected, the performance of all models generally deteriorates in the presence of noise; however, the degradation patterns vary across methods. The restored results from Dual-UNet \cite{25} collapse when heavy noise is added to both LR-HSI and HR-MSI. The other three methods are also affected to varying degrees. In contrast, our approach is notably resilient, maintaining higher PSNR and SAM values whether noise is added solely to LR-HSI or to both LR-HSI and HR-MSI.

The effectiveness of our proposed CSAKD framework in handling noise can be attributed to two key factors. First, the Cross Self-Attention (CSA) fusion module adaptively determines the optimal weights for features extracted from the LR-HSI and HR-MSI branches. By dynamically adjusting these weights based on the input data, the CSA module can effectively suppress the influence of noise and prioritize the more reliable information from each modality. Second, the Knowledge Distillation (KD) strategy enables the student network to learn robust feature representations from the teacher network. During training, the teacher network is exposed to noisy inputs and learns to extract noise-resilient features. Through the distillation process, this robustness is transferred to the student network, allowing it to maintain high-quality fusion results even in the presence of noise.

\begin{figure}[t!]
\centering
\includegraphics[width=1.03\linewidth]{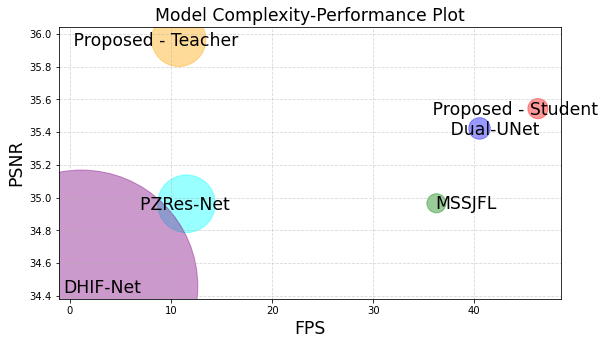}
\caption{Model complexity-performance comparison plot.The upper-right corner represents faster inference speed and higher fusion quality. The size of the circle means the memory-usage of the model deployed on hardware.}
\label{fig:cpplot}
\end{figure}
\subsection{Computational Complexity Analysis}
In addition to the restoration performance and robustness of the models, we also consider their computational complexity and lightweight nature for deployment on different hardware platforms. This is crucial because HSI processing needs to be compatible with various hardware constraints and SDG requirements, allowing for feasible applications without heavy computational burden. The results of the complexity analysis are shown in the last two columns of Table \ref{tab:performance} and Figure \ref{fig:cpplot}.

The HR-HSI restored from the state-of-the-art supervised learning-based methods all exhibit high fidelity. However, the model complexity and hardware requirements may vary in different aspects. The proposed method demonstrates comprehensively competitive capabilities across parameter size, FLOPs, running time per input pair, and memory usage during inference, underscoring an optimized balance between computational efficiency and fusion quality. By employing the proposed knowledge distillation framework, we significantly reduce the model size, FLOPs, and memory usage, which is extremely valuable for lightweight hardware platforms.

The other methods face different challenges arising from their drawbacks in handling heavy noise or their substantial hardware requirements. DHIF-Net \cite{26} and PZRes-Net \cite{24} are limited in their applicability to lightweight hardware due to their iterative spatial-spectral-aware optimization strategy or residual learning-based approach, which result in heavy parameter and memory requirements. Dual-UNet \cite{25} achieves low computational complexity but struggles to address highly noisy data. MSSJFL \cite{23} strikes a balance between maintaining fusion quality in the presence of noise and computational complexity, but its performance is relatively limited compared to our method.

\begin{figure}[t!]
\centering
\includegraphics[width=1\linewidth]{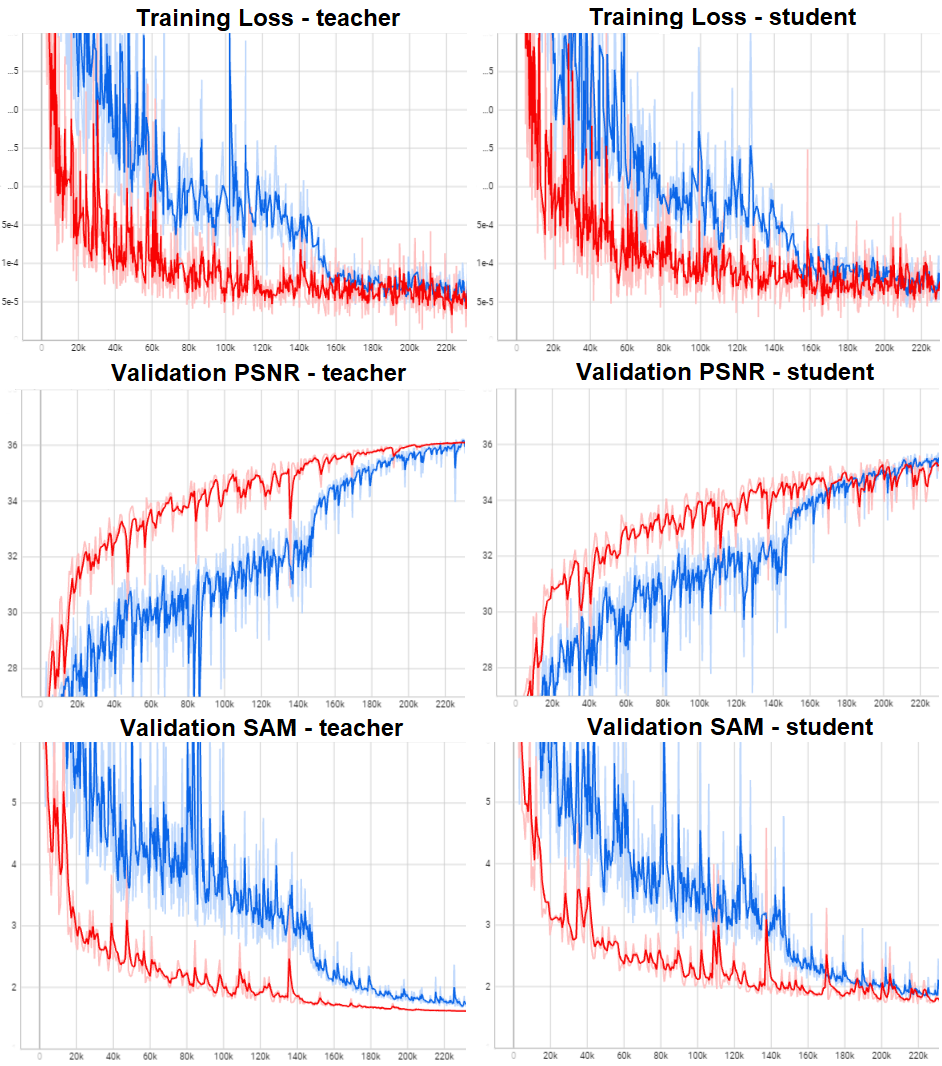}
\caption{The convergence process at the training phase and the validation performance comparison against the proposed loss function is employed or not. The red curve represents the proposed loss function is used. The blue curve denotes using the naive loss function, which means we just uses L1-loss in teacher model, L1-loss and response distillation loss term to guide lightweight student network.}
\label{fig:loss}
\end{figure}

\subsection{Model Scalability and Extended Training}
\begin{figure}
\centering
\includegraphics[width=1\linewidth]{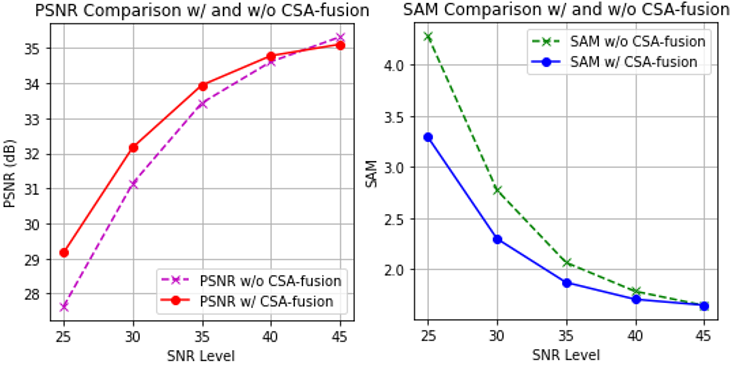}
\caption{Performance comparison between with and without proposed CSA-fusion block under different SNR level. Compared with direct fusion, the CSA-fusion method we proposed is more robust and effective against noise.}
\label{fig:csakd_ablation}
\end{figure}

\begin{table}[]
\centering
\caption{Comparisons of Different Coefficients of Penalty Term in the Proposed Student Model. The proposed setting means that the all coefficient of penalty term is set to 0.1. The naive setting means prohibiting the SAM, BEBA, feature map-KD loss.}
\label{tab:lossablation}
\scalebox{0.94}{
\begin{tabular}{c|c|c|c}
Teacher & PSNR / SAM / RMSE & Student & PSNR / SAM / RMSE \\ \hline
Naive & 20.430 / 8.145 / 148.33 & 
Naive & 19.266 / 8.448 / 171.361 \\  
$\lambda_1$=0.5 & 35.794 / 1.565 / 31.514 & $\lambda_1$=0.5 & 35.021 / 1.732 / 33.899 \\
$\lambda_2$=0.5 & 35.812 / 1.561 / 31.475 & $\lambda_2$=0.5 & 34.906 / 1.775 / 34.532 \\
$\lambda_3$=0.5 & 35.785 / 1.565 / 31.513 & $\lambda_3$=0.5 & 35.041 / 1.717 / 33.638 \\
$\lambda_4$=0.5 & 35.781 / 1.566 / 31.552 & $\lambda_4$=0.5 & 35.016 / 1.709 / 33.358 \\ \hline
$\textbf{Proposed}$ & 35.967 / 1.527 / 30.928 & $\textbf{Proposed}$ & 35.544 / 1.643 / 32.308
\end{tabular}}
\end{table}

The proposed CSAKD framework demonstrates exceptional performance, with both the teacher and student networks surpassing other state-of-the-art models. To explore the limitations of the CSAKD framework, which primarily depend on the DTS network architecture and the CSA-fusion module for achieving high-quality fusion results, we enhanced both networks by incorporating additional CLRA units. This enhancement aimed to assess their scalability and identify the upper bound of fusion performance, as detailed in Table \ref{tab:performance}.

Specifically, we augmented the teacher network by stacking the CLRA unit in the four branches ($\bZ_{h}, \bZ_{hm}, \bZ_{mh}, \bZ_{m}$) 8, 8, 6, and 6 times, respectively. In contrast, the student network received a more modest increase of 2, 4, 4, and 4 stacks. These versions, denoted as Proposed-Teacher-L and Proposed-Student-L, are presented in Table \ref{tab:performance}. This strategy served two purposes: first, to preserve the lightweight nature of the student network, and second, to amplify the learning capacity of the teacher model. The results indicate a significant improvement in the SAM and RMSE metrics for the student network. However, a limitation was observed in the PSNR metric. These findings suggest that our approach is viable for achieving superior fusion results when computational complexity is not a primary concern.

To further enhance the learning ability of the CSA-Large model, we explored the potential of a deeper teacher model, which can provide richer feature information in the feature domain. To investigate the effectiveness of the KD-guided framework, we extended the training process for CSA-Large by an additional 40 epochs and reduced the learning rate to $5\times10^{-5}$. These versions, denoted as Proposed-Teacher-L-Ext and Proposed-Student-L-Ext, are presented in Table \ref{tab:performance}. The results show that extending the training process degraded the teacher's performance, indicating that the model had reached its bottleneck. Conversely, the student model acquired more effective guidance, aligning its performance with the proposed KD framework. This demonstrates that the proposed feature-map knowledge distillation loss $\ell_{\text{KD}}$ can effectively enhance the student network when using a deeper teacher network and a richer feature space.

The scalability analysis highlights the flexibility of the CSAKD framework in accommodating varying network depths and architectures. By increasing the number of CLRA units, the fusion performance can be further improved, particularly in terms of the SAM and RMSE metrics. However, the limitations observed in the PSNR metric suggest that there may be a trade-off between network depth and certain aspects of fusion quality. This trade-off should be carefully considered when designing the network architecture for specific applications.

The extended training experiment demonstrates the effectiveness of the KD-guided framework in transferring knowledge from a deeper teacher network to a lightweight student network. By leveraging the richer feature space provided by the deeper teacher, the student network can learn more nuanced representations and achieve improved fusion performance. This finding underscores the importance of the feature-map knowledge distillation loss $\ell_{\text{KD}}$ in enabling effective knowledge transfer and enhancing the student network's learning ability.

In summary, the model scalability and extended training analysis provide valuable insights into the flexibility and effectiveness of the CSAKD framework. These findings can guide future research in designing and optimizing network architectures for HSI/MSI fusion tasks, while also highlighting the potential for further performance improvements through extended training and knowledge distillation.

\subsection{Ablation Study}

Based on Table \ref{tab:performance}, we corroborated the superior performance of the student network under teacher guidance. Furthermore, to explore the impact of teacher model complexity on student learning, highlighting the need for distillation loss to bridge the output gap between teacher and student, we perform the ablation study on loss function analysis shown as follows. In addition, we also perform the ablation study on CLRA depth analysis to determine the optimal number of CLRA units to balance the SR performance and the computational complexity, shown as follows. 




\subsubsection{Loss Function Analysis}
We first verify that the proposed loss function for joint training of the teacher-student model not only accelerates the training process, helping the model to converge at a high speed, but also effectively stabilizes the instability during backpropagation. As shown in Figure \ref{fig:loss} and Table \ref{tab:lossablation}, the proposed SAM loss and BEBA loss are both crucial for training a strong teacher model. Subsequently, the feature-map distillation loss enables the teacher model to guide the student model in an ideal manner.

Due to the complexity of the loss function in the student network, we compared the influence of each loss function component. The penalty term in the student network's loss function $\ell_{\text{total}}^{\text{student}}$ is calculated solely based on the discrepancy between the teacher-student network output and the ground truth, with the combined impact detailed in Table \ref{tab:lossablation}. In this experiment, the penalty terms in $\ell_{\text{total}}^{\text{teacher}}$ are all set to 0.1. The experiment demonstrates that the SAM loss is relatively sensitive in the CSAKD framework.

\subsubsection{CLRA Depth Analysis}
\begin{table}[t!]
\centering
\caption{Comparisons of Stacking Different Amounts of CLRA in Different Branches of the Proposed Model. \\M and G indicate $10^6$ and $10^9$.}
\label{tab:stackingCLRB}
\scalebox{1}{
\begin{tabular}{c|c|c|c}

$Z_{h}$,$Z_{hm}$,$Z_{mh}$,$Z_{m}$ &$\text{PSNR}$ / \text{SAM} / \text{RMSE} &FLOPs&Params\\ \hline
1, 3, 3, 3 & 35.528 / \best{1.598} / \best{31.698}  & 309G & 8.717M \\ 
2, 2, 2, 2 & 35.476 / 1.617 / 31.983  & 218G & \best{6.089M} \\ 
2, 3, 3, 2 & 35.405 / 1.624 / 32.127  & 224G & 7.418M \\\hline
\textbf{1, 4, 4, 1 (proposed)} & \best{35.544} / 1.643 / 32.308 & \best{144G} & 7.449M \\ 
\end{tabular}}
\end{table}    

In addition to comparing the computational complexity with other methods, we conducted experiments to determine the optimal depth combination of the CLRA units. The objective was to achieve the best balance between performance and speed. Table \ref{tab:stackingCLRB} presents the results of our experiments.

The depth of the CLRA units plays a crucial role in the fusion performance and computational efficiency of the proposed CSAKD framework. By varying the number of CLRA units in each branch of the DTS network, we can fine-tune the network's capacity to extract and integrate spatial-spectral features. The results in Table \ref{tab:stackingCLRB} demonstrate that the optimal combination of CLRA depths varies depending on the specific performance metrics and computational constraints.

For instance, the combination of 1, 3, 3, and 3 CLRA units in the $\bZ_h$, $\bZ_{hm}$, $\bZ_{mh}$, and $\bZ_m$ branches, respectively, achieves the best SAM and RMSE metrics. However, this configuration also results in a higher number of parameters and FLOPs compared to the proposed combination of 1, 4, 4, and 1 CLRA units. The proposed combination strikes a balance between performance and computational efficiency, achieving competitive PSNR and SAM metrics while maintaining a lower number of parameters and FLOPs.

These findings highlight the importance of carefully designing the network architecture and selecting the appropriate depth of the CLRA units based on the specific requirements of the application. The ablation study provides valuable insights into the trade-offs between fusion performance and computational complexity, enabling researchers and practitioners to make informed decisions when deploying the CSAKD framework in real-world scenarios.

\section{Conclusion}\label{sec:Conclusion}
In this work, we have introduced a novel knowledge distillation-based teacher-student framework, named CSAKD, for LR-HSI/HR-MSI fusion. The proposed framework incorporates a Dual Two-Streamed (DTS) network architecture, which effectively captures spectral and spatial information from LR-HSI and HR-MSI. The Cross-Layer Residual Aggregation (CLRA) unit and Cross Self-Attention (CSA) module enhance the network's ability to handle noise and integrate spatial-spectral features, resulting in high-quality fused results.
The application of knowledge distillation in LR-HSI/HR-MSI fusion is a key contribution of this work. The proposed Spectral Angle Mapper (SAM) loss, Band-Energy-Balance-Aware (BEBA) loss, and feature map-based KD loss guide the lightweight student model to achieve excellent fusion performance while reducing model-size and computational requirements.
Extensive experiments have demonstrated the superiority of the CSAKD method under various conditions, including noisy images and LR-HSIs with varying numbers of bands. The lightweight student model exhibits outstanding performance compared to larger, state-of-the-art models, offering an exceptional balance of high performance and reduced computational complexity.
The CSAKD framework opens up new possibilities for efficient and effective HSI/MSI fusion, with potential applications in remote sensing and related fields. Future research could explore integrating CSAKD with other advanced techniques, such as attention mechanisms and adversarial learning, to further improve fusion performance and adaptability to diverse scenarios.

\bibliographystyle{IEEEtran}
\bibliography{ref}

\end{document}